\documentclass[conference]{IEEEtran}
\IEEEoverridecommandlockouts
\usepackage{cite}
\usepackage{amsmath,amssymb,amsfonts}
\usepackage{algorithmic}
\usepackage{graphicx}
\usepackage{textcomp}
\usepackage{xcolor}
\usepackage{booktabs}
\usepackage{comment}

\def\BibTeX{{\rm B\kern-.05em{\sc i\kern-.025em b}\kern-.08em
    T\kern-.1667em\lower.7ex\hbox{E}\kern-.125emX}}
\begin{document}

\title{Refusal Behavior in Large Language Models: \\ A Nonlinear Perspective}

\author{
\IEEEauthorblockN{Fabian Hildebrandt}
\IEEEauthorblockA{
\textit{CCN Group} \\
\textit{Pattern Recog. Lab.} \\
\textit{FAU Erlangen-Nürnberg}\\
Erlangen, Germany \\
fabian.hildebrandt@fau.de}
\and
\IEEEauthorblockN{Andreas Maier}
\IEEEauthorblockA{
\textit{Pattern Recog. Lab.} \\
\textit{FAU Erlangen-Nürnberg}\\
Erlangen, Germany \\
andreas.maier@fau.de\\
}
\and
\IEEEauthorblockN{Patrick Krauss$^{*}$}
\IEEEauthorblockA{
\textit{CCN Group} \\
\textit{Pattern Recog. Lab.} \\
\textit{FAU Erlangen-Nürnberg}\\
Erlangen, Germany \\
patrick.krauss@fau.de}
\and
\IEEEauthorblockN{Achim Schilling$^{*}$}
\IEEEauthorblockA{
\textit{Neuroscience Lab,} \\
\textit{University Hospital Erlangen,}\\ 
\textit{CCN Group} \\
\textit{Pattern Recog. Lab.} \\
\textit{FAU Erlangen-Nürnberg}\\
Erlangen, Germany \\
achim.schilling@fau.de
}
}


\maketitle

\begin{abstract}
Refusal behavior in large language models (LLMs) enables them to decline responding to harmful, unethical, or inappropriate prompts, ensuring alignment with ethical standards. This paper investigates refusal behavior across six LLMs from three architectural families. We challenge the assumption of refusal as a linear phenomenon by employing dimensionality reduction techniques, including PCA, t-SNE, and UMAP. Our results reveal that refusal mechanisms exhibit nonlinear, multidimensional characteristics that vary by model architecture and layer. These findings highlight the need for nonlinear interpretability to improve alignment research and inform safer AI deployment strategies.
\end{abstract}

\begin{IEEEkeywords}
refusal, mechanistic interpretability, LLM, AI alignment, neuroscience-inspired AI (neuroAI), explainable AI (XAI)
\end{IEEEkeywords}

\section*{Introduction}
Morality and impulse control are fundamental aspects of human behavior, enabling ethical decision-making, effective social interactions, and the maintenance of personal and societal relationships. The amygdala and the ventromedial prefrontal cortex play critical roles in moral behavior \cite{blair2007amygdala_vmPFC}. The amygdala associates harmful actions with negative emotions, while the ventromedial prefrontal cortex governs decision-making and self-control to prevent such actions. In psychopathy, these regions exhibit reduced activity, leading to an increased propensity for manipulation, impulsivity, and unethical decisions without remorse \cite{glenn2009moral_psychopathy}. Impulse control, a key aspect of self-regulation housed in the prefrontal cortex, can be disrupted by various factors such as ADHD, substance use disorders, psychological conditions, stress, and brain injuries \cite{kim2011prefrontal}.

The human brain is often described as a prediction machine, continuously processing sensory information to guide decision-making. Similarly, Large Language Models (LLMs) are designed to predict the next word or sequence, enabling them to perform complex language tasks. Like the human brain, LLMs are tasked with handling nuanced decisions, such as refusing harmful instructions, yet their internal decision-making processes often remain opaque. Efforts to align LLMs with ethical principles have focused on feature search, which seeks to interpret model activations and uncover the mechanisms governing their behavior. The superposition hypothesis posits that multiple concepts or features are simultaneously represented within the same neural activation space \cite{olah2020zoom, elhage2022toymodelssuperposition}. Consequently, individual neurons in LLMs often exhibit polysemantic behavior, encoding multiple distinct meanings rather than a one-to-one mapping. Achieving monosemantic representations—where neurons encode a single, unambiguous feature—is a critical goal for improving interpretability. Sparse autoencoders have been employed to identify such features by inflating the hidden space and enforcing sparsity \cite{bricken2023monosemanticity, faruqui2015sparseovercompletewordvector, goh2016thoughtvector, arora2018linearalgebraicstructureword}. This approach enables the isolation of specific features for steering model behavior \cite{zou2023representationengineeringtopdownapproach, turner2024steeringlanguagemodelsactivation, panickssery2024steeringllama2contrastive}.

Recent studies suggest that refusal behavior in LLMs is mediated by a single linear subspace in the activation space. This feature can be manipulated to either disable or enforce refusal behavior across a range of open-source models, including those with up to 72B parameters \cite{arditi2024refusallanguagemodelsmediated}. Techniques like the difference-in-means method \cite{belrose2023diff} and weight orthogonalization have been used to isolate and modify this subspace, resulting in models that either lose or gain the ability to refuse harmful instructions. However, emerging evidence challenges the assumption that refusal behavior resides in a linear subspace, pointing instead to a multidimensional and nonlinear nature \cite{engels2024languagemodelfeatureslinear}.

In this work, we examine refusal behavior across six LLMs spanning three model families. By analyzing intermediate layer activations through both linear (PCA) and nonlinear (t-SNE, UMAP) dimensionality reduction techniques, we reveal that refusal behavior is a universal but architecture-specific phenomenon. Our findings show that refusal mechanisms are more complex and nonlinear than previously assumed, with distinct sub-clusters emerging in the activation space. These insights contribute to a deeper understanding of refusal behavior and its implications for aligning LLMs with ethical and safety standards.

\section*{Methodology}

\subsection*{Behavioral studies and neural correlates}

In this study, we investigate the refusal behavior of large language models (LLMs) in response to harmful and harmless instructions, drawing parallels to behavioral studies that examine brain activity linked to cognitive functions. Our methodology involves designing a task that isolates the refusal mechanisms in LLMs, similar to decision-making tasks in behavioral research that require empathy and sensitivity. Simultaneously, we track the LLM activations to identify model activations associated with the refusal behavior. Additionally, the origin of these responses is localized and the embeddings are further investigated.

\subsection*{Datasets}

Two distinct datasets are used. The first dataset $D_{\text{harmless}}$ contains harmless instructions from the ALPACA dataset \cite{taori2023stanford} repackaged and published as a Hugging Face dataset \cite{labonne_harmless_alpaca}. The ALPACA dataset is a collection of 52000 harmless instruction-following prompts designed to fine-tune large language models for more effective and reliable task completion. The second dataset $D_{\text{harmful}}$ contains harmful instructions that originate from the \textit{LLM Attacks} dataset \cite{zou2023universal}, which was designed for adversarial attacks against aligned language models. The instructions are trying to induce harmful behavior by the LLMs. Again, a repackaged Hugging Face dataset is used \cite{labonne_harmful_behaviors}.

\subsection*{Models}

To investigate the universality of the refusal behavior, we evaluated a set of six large language models (LLMs) from three model families. All models are fine-tuned for instruction following. The selection of different model families and varying parameter sizes allows to assess whether the observed effects are consistent and generalizable. The model families represent different alignment types from preference optimization to alignment by fine-tuning. The specific models used in this study are detailed in Table \ref{tab:model_comparison}.

\begin{table*}[h!]
\centering
\caption{Comparison of Model Specifications}
\label{tab:model_comparison}
\resizebox{\textwidth}{!}{%
\begin{tabular}{lcccl}
\hline
\textbf{Model Name} & \textbf{Model Size} & \textbf{Number of Layers} & \textbf{Hidden Space Dimensions} & \textbf{Model Origin} \\ \hline
Llama-3.2-1B-Instruct & 1.1B & 16 & 2048 & Meta AI \cite{meta_llama_3.2} \\ 
Llama-3.2-3B-Instruct & 3.2B & 28 & 3072 & Meta AI \cite{meta_llama_3.2}\\ 
Bloom-560M-Instruct & 302M & 24 & 1024 & BigScience \cite{workshop2023bloom176bparameteropenaccessmultilingual} \\ 
Bloom-3B-Instruct & 2.4B & 30 & 2560 & BigScience \cite{workshop2023bloom176bparameteropenaccessmultilingual} \\ 
Qwen2-0.5B-Instruct & 391M & 24 & 896 & Alibaba Group \cite{yang2024qwen2technicalreport} \\ 
Qwen2-1.5B-Instruct & 1.4B & 28 & 1536 & Alibaba Group \cite{yang2024qwen2technicalreport} \\ \hline
\end{tabular}%
}
\end{table*}

\subsection*{Extraction of the refusal activations}

The methodology for extracting activations related to the refusal mechanism follows the general approach outlined in \cite{arditi2024refusallanguagemodelsmediated} and \cite{labonne2024uncensor}. First, we load the model using the \texttt{TranformerLens} library \cite{nanda2022transformerlens}. The library allows to cache any internal activation in the model for mechanistic interpretability research. Next, two equal-sized subsets of prompts from the two datasets $D_{\text{harmful}}$ and $D_{\text{harmless}}$ containing harmful and harmless instructions are loaded using the Hugging Face \texttt{Datasets} library. Using a chat-style generation template, we run the model on both datasets while caching the activations for each layer. To identify the refusal-related feature direction, we compute the difference-in-means by subtracting the mean activations of harmless prompts from those of harmful prompts at each token position. Finally, we store the residual activations afterr the self-attention layers and the multilayer perceptron (MLP) layers  at the last token position, for further analysis and manipulation.

\textbf{Difference-in-means.} The difference-in-means technique \cite{belrose2023diff} can be used to identify the refusal direction in the model's activations as shown in previous works \cite{arditi2024refusallanguagemodelsmediated}, \cite{labonne2024uncensor}. 

For each layer $l \in [L]$ the mean activation $\mu_\text{harmful}^{(l)}$ for harmful prompts from $D_{\text{harmful}}$ and $\mu_\text{harmless}^{(l)}$ for harmless prompts from $D_{\text{harmless}}$ is calculated:

\begin{equation}
\begin{split}
\mu_\text{harmful}^{(l)} &= \frac{1}{|D_{\text{harmful}}|} \sum_{i \in D_{\text{harmful}}} y^{(l)}(i), \\
\mu_\text{harmless}^{(l)} &= \frac{1}{|D_{\text{harmless}}|} \sum_{i \in D_{\text{harmless}}} y^{(l)}(i).
\end{split}
\end{equation}

where $y_i^{(l)}(t)$ represents the activation at the last token position in layer $l$ for instruction $i$. The difference-in-means direction $d_\text{refusal}^{(l)}$ is then computed as:

\begin{equation}
d_\text{refusal}^{(l)}= \mu_\text{harmfull}^{(l)} - \mu_\text{harmless}^{(l)}.
\end{equation}

The vector $d_\text{refusal}^{(l)}$ captures both the direction and the magnitude of the refusal feature in the activations. 

\subsection*{Dimensionality reduction}

\textbf{Principal Component Analysis (PCA).} PCA is a widely used linear dimensionality reduction technique that transforms an unlabeled dataset into a new coordinate system where the greatest variance by any projection of the data lies on the first principal component, the second greatest variance on the second component, and so on \cite{pearson1901pca}. This method identifies orthogonal axes that maximize variance, thereby simplifying high-dimensional data while preserving the maximum variability. PCA is commonly applied for feature extraction and the feature visualization.

\textbf{t-Distributed Stochastic Neighbor Embedding (t-SNE).} t-SNE is a nonlinear dimensionality reduction technique commonly used for visualizing high-dimensional data in two or three dimensions. It models the pairwise similarities between points in the original space and seeks to preserve these relationships in the reduced space by minimizing a divergence between two probability distributions. Unlike linear methods like PCA, t-SNE is effective at capturing local structure and revealing clusters or patterns within complex datasets \cite{van2008tsne}.

\textbf{Uniform Manifold Approximation and Projection (UMAP).}
UMAP is a dimensionality reduction technique designed for visualization and the general nonlinear embedding of high-dimensional data. Based on manifold learning and topological data analysis, UMAP constructs a high-dimensional graph of data points and optimizes a low-dimensional projection to preserve both local and global structures. Compared to PCA, UMAP captures complex, nonlinear relationships in data, making it more effective for revealing meaningful clusters. Unlike t-SNE, UMAP offers faster computation, better scalability, and more interpretable low-dimensional embeddings while maintaining the ability to preserve local neighborhood information \cite{mcinnes2018umap}. 

\subsection*{Activation separability metric}

\textbf{Generalized Discrimination Value (GDV).} To quantify the degree of clustering, we used the GDV as published and explained in detail in \cite{schilling2021gdv}. The GDV provides an objective measure of how well the hidden layer activations cluster according to the ASC types, offering insights into the model's internal representations. Briefly, we consider $N$ points $\mathbf{x_{n=1..N}}=(x_{n,1},\cdots,x_{n,D})$, distributed within $D$-dimensional space. A label $l_n$ assigns each point to one of $L$ distinct classes $C_{l=1..L}$. In order to become invariant against scaling and translation, each dimension is separately z-scored and, for later convenience, multiplied with $\frac{1}{2}$:
\begin{align}
s_{n,d}=\frac{1}{2}\cdot\frac{x_{n,d}-\mu_d}{\sigma_d}.
\end{align}
Here, $\mu_d=\frac{1}{N}\sum_{n=1}^{N}x_{n,d}\;$ denotes the mean,\\ \\
and $\sigma_d=\sqrt{\frac{1}{N}\sum_{n=1}^{N}(x_{n,d}-\mu_d)^2}$ the standard deviation of dimension $d$. \\ \\
Based on the re-scaled data points $\mathbf{s_n}=(s_{n,1},\cdots,s_{n,D})$, we calculate the {\em mean intra-class distances} for each class $C_l$ 
\begin{align}
\bar{d}(C_l)=\frac{2}{N_l (N_l\!-\!1)}\sum_{i=1}^{N_l-1}\sum_{j=i+1}^{N_l}{d(\textbf{s}_{i}^{(l)},\textbf{s}_{j}^{(l)})},
\end{align}
and the {\em mean inter-class distances} for each pair of classes $C_l$ and $C_m$
\begin{align}
\bar{d}(C_l,C_m)=\frac{1}{N_l  N_m}\sum_{i=1}^{N_l}\sum_{j=1}^{N_m}{d(\textbf{s}_{i}^{(l)},\textbf{s}_{j}^{(m)})}.
\end{align}
Here, $N_k$ is the number of points in class $k$, and $\textbf{s}_{i}^{(k)}$ is the $i^{th}$ point of class $k$.
The quantity $d(\textbf{a},\textbf{b})$ is the euclidean distance between $\textbf{a}$ and $\textbf{b}$. Finally, the Generalized Discrimination Value (GDV) is calculated from the mean intra-class and inter-class distances  as follows:
\begin{align}
\mbox{GDV}=\frac{1}{\sqrt{D}}\left[\frac{1}{L}\sum_{l=1}^L{\bar{d}(C_l)}\;-\;\frac{2}{L(L\!-\!1)}\sum_{l=1}^{L-1}\sum_{m=l+1}^{L}\bar{d}(C_l,C_m)\right]
 \label{GDVEq}
\end{align}

\noindent whereas the factor $\frac{1}{\sqrt{D}}$ is introduced for dimensionality invariance of the GDV with $D$ as the number of dimensions.

\vspace{0.2cm}\noindent Note that the GDV is invariant with respect to a global scaling or shifting of the data (due to the z-scoring), and also invariant with respect to a permutation of the components in the $N$-dimensional data vectors (because the euclidean distance measure has this symmetry). The GDV is zero for completely overlapping, non-separated clusters, and it becomes more negative as the separation increases. A GDV of -1 signifies already a very strong separation and perfect clustering.  

\section*{Results}

\subsection*{Refusal is a universal feature}

\begin{figure*}[htbp]
\centerline{\includegraphics[width=0.6\textwidth]{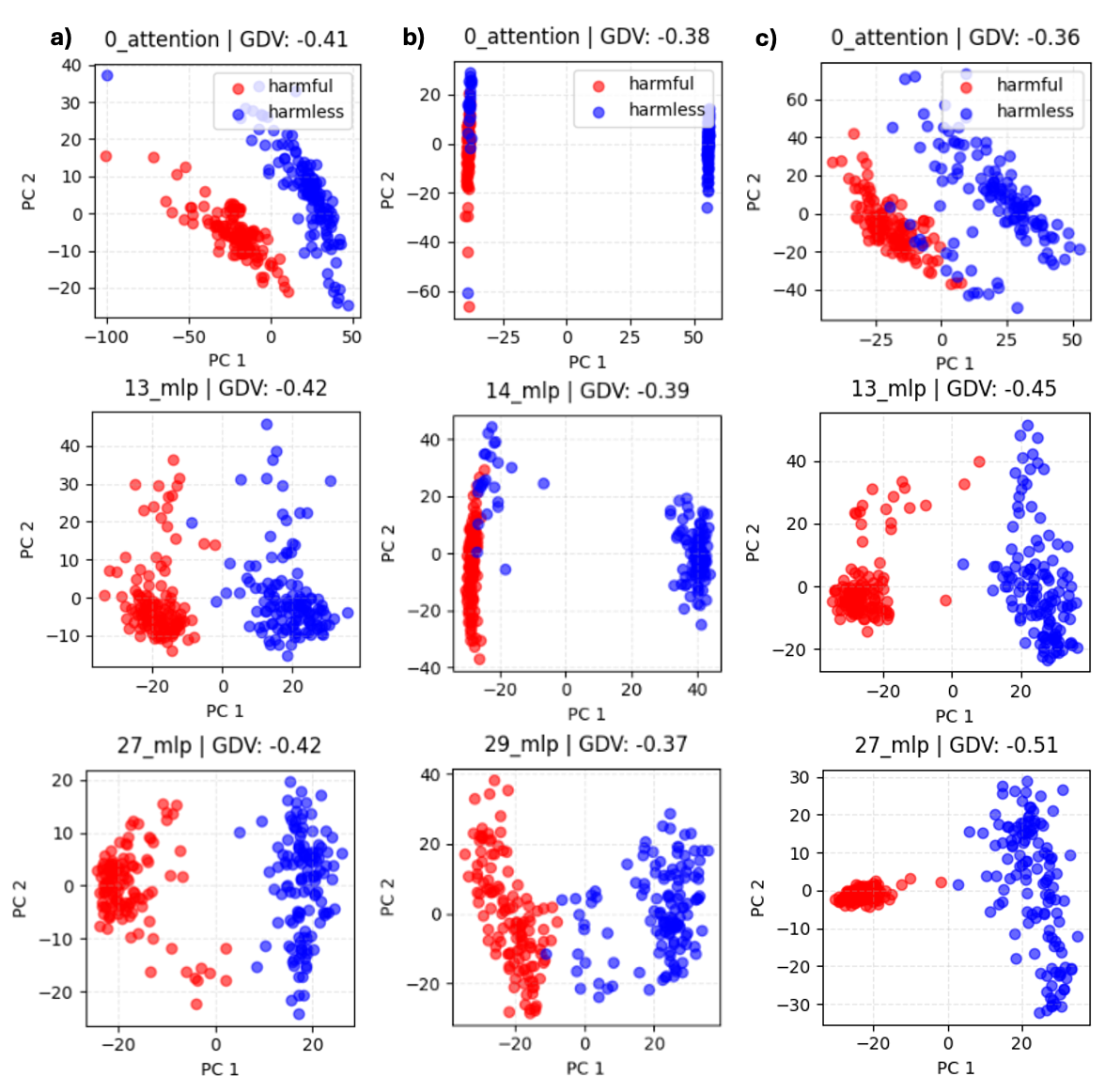}}
\caption{Dimensionality-reduced residual activations using PCA, comparing harmful and harmless instructions across three different models and layers. \textbf{a)} Qwen2-1.5B-Instruct model (first, middle, last layers). \textbf{b)} Bloom-3b model (first, middle, last layers). \textbf{c)} Llama-3.2-3B-Instruct model (first, middle, last layers).}
\label{fig:model_scatter_comparison}
\end{figure*}

Refusal behavior is consistently observed across all six tested models, with a distinct separation between harmful and harmless instructions in the dimensionality-reduced residual activations. This separation is evident across model families and sizes. Figure \ref{fig:model_scatter_comparison} illustrates the refusal feature in three models at different layers. Table \ref{tab:model_gdv_comparison} summarizes the lowest GDV values, indicating the best separability between harmful and harmless instructions, along with the corresponding model layers and dimensionality-reduction techniques.

\begin{table}[htbp]
\centering
\caption{Comparison of Separability across Models and Methods}
\label{tab:model_gdv_comparison}
\begin{tabular}{llrl}
\toprule
\textbf{Model} & \textbf{Method} & \textbf{GDV} & \textbf{Layer} \\
\midrule
Llama-3.2-1B-Instruct & PCA & -0.45 & 12/16 (ATTENTION) \\
Llama-3.2-1B-Instruct & t-SNE & -0.62 & 9/16 (MLP) \\
Llama-3.2-1B-Instruct & UMAP & -0.69 & 16/16 (MLP) \\
Llama-3.2-3B-Instruct & PCA & -0.51 & 24/28 (ATTENTION) \\
Llama-3.2-3B-Instruct & t-SNE & -0.60 & 11/28 (MLP) \\
Llama-3.2-3B-Instruct & UMAP & -0.80 & 15/28 (ATTENTION) \\
Qwen2-0.5B-Instruct & PCA & -0.46 & 1/24 (ATTENTION) \\
Qwen2-0.5B-Instruct & t-SNE & -0.73 & 5/24 (ATTENTION) \\
Qwen2-0.5B-Instruct & UMAP & -0.87 & 5/24 (MLP) \\
Qwen2-1.5B-Instruct & PCA & -0.44 & 4/28 (ATTENTION) \\
Qwen2-1.5B-Instruct & t-SNE & -0.71 & 2/28 (MLP) \\
Qwen2-1.5B-Instruct & UMAP & -0.89 & 18/28 (ATTENTION) \\
bloom-3b & PCA & -0.40 & 6/30 (MLP) \\
bloom-3b & t-SNE & -0.63 & 28/30 (ATTENTION) \\
bloom-3b & UMAP & -0.66 & 25/30 (MLP) \\
bloom-560m & PCA & -0.41 & 5/24 (MLP) \\
bloom-560m & t-SNE & -0.60 & 23/24 (ATTENTION) \\
bloom-560m & UMAP & -0.65 & 22/24 (MLP) \\
\bottomrule
\end{tabular}
\end{table}

\subsection*{Refusal is a nonlinear feature}

Our analysis demonstrates that refusal behavior in LLMs extends beyond a simple one-dimensional linear subspace, revealing a complex, nonlinear structure. Using PCA, UMAP, and t-SNE, we visualized dimensionality-reduced activations for harmful and harmless instructions across multiple layers of six LLMs. While PCA, a linear technique, captures variance through linear distances, UMAP and t-SNE effectively identify nonlinear relationships, offering deeper insights into the activation space.

\begin{figure}[htbp]
\centerline{\includegraphics[width=0.7\columnwidth]{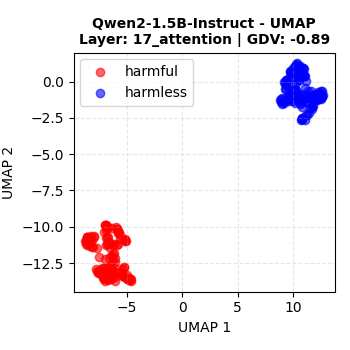}}
\caption{Dimensionality-reduced residual activations of the Qwen2-1.5B-Instruct model at layer 18, visualized using UMAP, showing distinct clusters for harmful and harmless instructions.}
\label{fig:compact_clusters}
\end{figure}

UMAP and t-SNE consistently yielded clearer separations than PCA, as indicated by lower GDV scores across all models. For instance, the Qwen2-1.5B-Instruct model achieved nearly perfect cluster separability in its 18th layer, with distinct, compact clusters for harmful and harmless instructions (Fig. \ref{fig:compact_clusters}). These clusters, characterized by large inter-cluster and compact intra-cluster distances, provide a strong basis for probing classifier training.

\begin{figure}[htbp]
\centerline{\includegraphics[width=0.7\columnwidth]{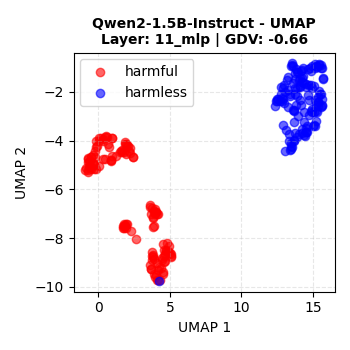}}
\caption{Dimensionality-reduced residual activations of the Qwen2-1.5B-Instruct model at layer 11, showing the emergence of sub-clusters for harmful instructions.}
\label{fig:qwen_subclusters}
\end{figure}

Further analysis revealed evolving cluster morphologies across layers. Linear separability changed gradually, while UMAP and t-SNE highlighted dynamic variations in cluster shapes and the emergence of sub-clusters. These sub-clusters, as seen in Fig. \ref{fig:qwen_subclusters}, suggest additional nuanced features, potentially indicating the division of harmful instructions into finer sub-features.

These findings underscore that refusal behavior is a multidimensional and nonlinear phenomenon, necessitating advanced techniques for comprehensive analysis and interpretability.

\subsection*{Distinct refusal mechanisms across different model families}

\begin{figure*}[htbp]
\centerline{\includegraphics[width=0.7\textwidth]{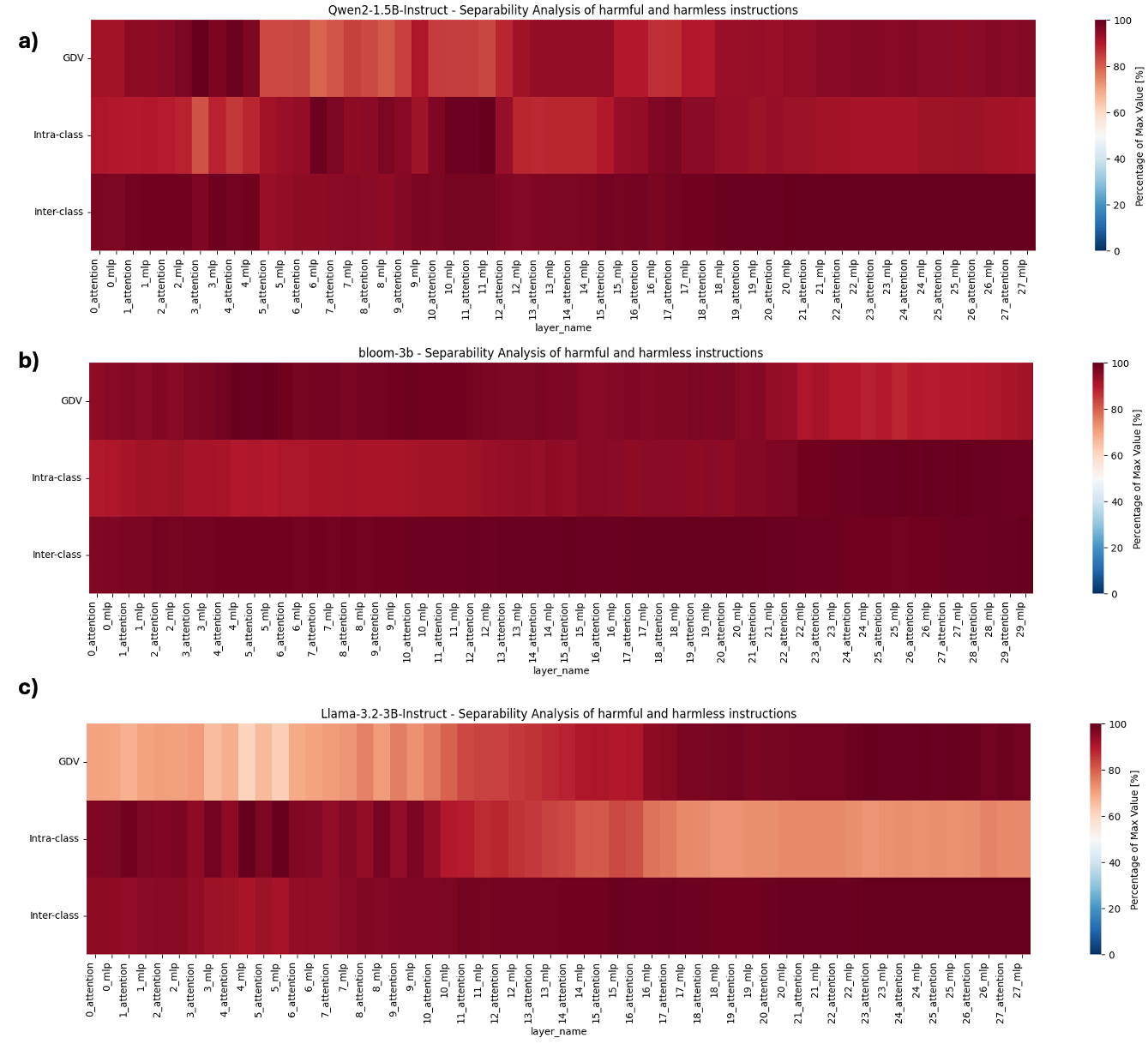}}
\caption{GDV, intra-class distance (compactness of harmful and harmless clusters), and inter-class distance (separation between clusters) of the dimensionality-reduced embeddings using PCA. \textbf{a)} Qwen2-1.5B-Instruct demonstrates early layer dominance of the refusal feature. \textbf{b)} Bloom-3b shows peak separability at early to intermediate layers but weaker discrimination in later layers. \textbf{c)} Llama-3.2-3B-Instruct displays progressively stronger separation.}
\label{fig:gdv_model_comparison}
\end{figure*}

Unlike the localized functional architecture of the human brain, where specialized areas handle specific tasks, LLMs exhibit diverse strategies for embedding the distinction between harmful and harmless instructions. Refusal behavior varies across model families and layers, as visualized in Fig. \ref{fig:gdv_model_comparison}.

\textbf{Qwen2 Models.} In the Qwen2 family, refusal behavior is primarily encoded in early layers. PCA reveals that the refusal feature emerges within the first few layers, with the 0.5B model peaking at the first layer and the 1.5B model at the fourth. Early layers integrate multiple principal components to define the refusal direction, while later layers refine it along the first principal component. This progression leads to stable inter-class distances and declining intra-class distances, resulting in compact harmful clusters. UMAP and t-SNE further confirm robust separability after the fifth layer.

\textbf{Bloom Models.} The Bloom architecture exhibits a distinct refusal mechanism, with a tendency to misclassify harmless instructions as harmful. The refusal direction aligns with the first principal component or main UMAP/t-SNE direction from the initial layer, where instructions are classified in a near-binary manner. GDV peaks in the fifth and sixth layers, but subsequent layers show weakened discrimination as intra-class distances increase and inter-class distances decrease. UMAP visualizations reveal sub-clusters within harmless instructions.

\textbf{Llama Models.} Llama models display a gradual refinement of refusal behavior. In early layers, harmful and harmless embeddings are mixed, indicating weak differentiation. Refusal strength intensifies in deeper layers, with Llama-3.2-3B-Instruct achieving compact clusters for harmful instructions and widespread embeddings for harmless ones, demonstrating stronger recognition. Conversely, the smaller Llama-3.2-1B-Instruct, with fewer hidden dimensions, shows more mixed embeddings, consistent with the superposition hypothesis. Nonlinear methods, such as t-SNE and UMAP, corroborate the emergence of clearer refusal representations in middle layers.

These findings reveal that refusal behavior is universally present but uniquely mediated across different architectures, reflecting varied strategies for harmful content differentiation.

\section*{Discussion}

Refusal behavior, enabling differentiation between harmful and harmless instructions, was consistently observed across all six models tested, corroborating previous findings \cite{arditi2024refusallanguagemodelsmediated}. 

Our analysis highlights that refusal mechanisms vary significantly across model families, reflecting architecture-specific strategies for harmful content detection. Qwen2 models primarily encode refusal features in early layers, achieving stable inter-class distances and compact intra-class structures. Bloom models exhibit peak separability in intermediate layers but demonstrate weaker discrimination in later layers, often misclassifying harmless instructions. Conversely, Llama models refine refusal gradually across layers, with stronger separability emerging in deeper layers.

Contrary to the assumption that refusal resides in a simple one-dimensional linear subspace \cite{arditi2024refusallanguagemodelsmediated}, our findings reveal a more complex, nonlinear structure. Using PCA, UMAP, and t-SNE, we visualized latent space activations and found that nonlinear methods consistently outperformed PCA in separating harmful and harmless embeddings, as evidenced by the separability metric GDV.

These results align with recent studies on jailbreak attacks, which indicate that nonlinear features, rather than universal linear ones, drive the success of such attacks \cite{kirch2024featurespromptsjailbreakllms}. Although focused on jailbreak mechanisms, these findings suggest a potential connection between jailbreak features and the refusal and harmfulness features, given the critical role of refusal in determining whether a model responds or declines.

Furthermore, the phenomenon of "alignment faking," where LLMs pretend to align with training objectives while resisting preference modifications \cite{greenblatt2024alignmentfakinglargelanguage}, raises concerns about the robustness of model alignment processes. Persistent alignment faking could lock in model preferences, posing challenges for future fine-tuning efforts.

These insights underscore the importance of nonlinear interpretability techniques for understanding refusal behavior and its implications for alignment, robustness, and ethical AI deployment.

\section*{Conclusion}

This study investigates refusal behavior in large language models (LLMs) and provides new insights into its mechanisms and characteristics. By analyzing six LLMs across three architectural families, we demonstrate that refusal behavior is a universal phenomenon but exhibits architecture-specific patterns. Contrary to prior assumptions that refusal mechanisms are linear and confined to single activation directions, our findings reveal that they are inherently nonlinear and multidimensional. Using advanced dimensionality reduction techniques such as UMAP and t-SNE, we uncover richer and more complex activation patterns than those detected through traditional linear methods like PCA.

The results emphasize the importance of nonlinear interpretability methods in understanding and improving model alignment. Refusal mechanisms vary significantly across architectures, with Qwen models encoding refusal early, Bloom models demonstrating intermediate-layer strengths, and Llama models refining refusal behavior in deeper layers. These differences highlight the diverse strategies employed by LLMs to distinguish harmful and harmless prompts, which has implications for designing safer, more robust AI systems.

Future research should explore the transferability of nonlinear refusal probes across model families and examine their interplay with jailbreak attacks and other adversarial strategies. Understanding refusal behavior's nuanced and nonlinear nature will be pivotal in enhancing the alignment, transparency, and ethical deployment of LLMs.

\section*{Author contributions}


FH, PK and AS conceptualized the study and designed the study. FH conducted the experiments. PK, AM and AS provided guidance, supervision, and feedback on the research design and interpretation of results. All authors wrote, reviewed and approved the final manuscript.

\section*{Code availability}


All results and the code are published on a public GitHub repository \cite{hildebrandt2024refusalllms} under the MIT license.

\section*{Acknowledgements}

This work was funded by the Deutsche Forschungsgemeinschaft (DFG, German Research Foundation): KR\,5148/3-1 (project number 510395418), KR\,5148/5-1 (project number 542747151), and GRK\,2839 (project number 468527017) to PK, and grant SCHI\,1482/3-1 (project number 451810794) to AS. 




\end{document}